# Spatial Pyramid Convolutional Neural Network for Social Event Detection in Static Image

Reza Fuad Rachmadi*, Keiichi Uchimura, and Gou Koutaki
Graduate School of Science and Technology, Kumamoto University

*Abstract-* Social event detection in a static image is a very challenging problem and it's very useful for internet of things applications including automatic photo organization, ads recommender system, or image captioning. Several publications show that variety of objects, scene, and people can be very ambiguous for the system to decide the event that occurs in the image. We proposed the spatial pyramid configuration of convolutional neural network (CNN) classifier for social event detection in a static image. By applying the spatial pyramid configuration to the CNN classifier, the detail that occurs in the image can observe more accurately by the classifier. USED dataset provided by Ahmad et al. is used to evaluate our proposed method, which consists of two different image sets, EiMM, and SED dataset. As a result, the average accuracy of our system outperforms the baseline method by 15% and 2% respectively.

## I. INTRODUCTION

Images and video is very popular data that uploaded to the internet, especially in the era of Internet of Things (IoT). The application to analyze the images and video is required to use the data in the meaningful ways, for the examples ads recommender system, regrouping the photo collections data, and images tagging application. There a lot of algorithm being proposed to tackle image understanding problem, including which described in [1,4,5]. Recently, the deep learning classifier is a very popular method to tackle a lot of problems and it's can deal with variant of the data, like images, video, or sound.

In the paper, we proposed a spatial pyramid convolutional neural network (CNN) classifier for classifying the event in the image. Spatial pyramid configuration of the image is used as input to the classifier and it can representing the detail of the image to the classifier. USED dataset described in [1] is used for our experiments.

## II. THE PROPOSED SYSTEM

We proposed a spatial pyramid CNN (SP-CNN) classifier to solve the event detection problem. The diagram of our spatial CNN classifier shows in Fig. 1. The spatial pyramid configuration is inspired by Lazebnik et al. [5], which represent the image by histogram for each spatial pyramid area and used as features for matching two images. The original Alex-Net CNN architecture is used for each stream of the classifier. Two levels of pyramid split are used to create the classifier with the total number of five streams of CNN classifier.

### A. Data Preparation

All images for training and testing process are resized to 256x256@3 resolution and the data is shifted to zero centered mean data by subtracting the data with the mean of the data. The split configuration of the dataset from the original dataset is used for the experiments. The USED dataset consists of two different datasets, including the EiMM dataset and SED dataset. The EiMM dataset consists of 216,000 examples with 160,000 examples for training process and 56,000 examples for testing process. The total number of classes in EiMM dataset is eight classes, including concert, graduation, meeting, mountain trip, picnic, sea holiday, ski holiday, and wedding class. The SED dataset consists of 189,000 examples with 140,000 examples for training process and 49,000 examples for testing process. The total number of classes in SED dataset is seven classes including concert, conference, exhibition, fashion, protest, sport, and theater/dance. The USED dataset is a multimedia dataset which also contains some video in the training or testing split configuration.

### B. The Training Process

The SED and EiMM subset of USED dataset are trained separately using the same classifier. The classifier uses the same parameters of the training process for two datasets. The training process was running for 32,000 iterations using Caffe framework [3] with stochastic gradient descent (SGD) algorithm and mini batch examples of 256. The learning rate α = 0.01 is used with momentum $m = 0.9$ and weight decay $\zeta = 0.0005$. The weights of the classifier are initialized using weights trained using ImageNet dataset [2] to reduce the overfitting problem. In the training process, only image data that listed as a training dataset and the video data is only used as testing process.

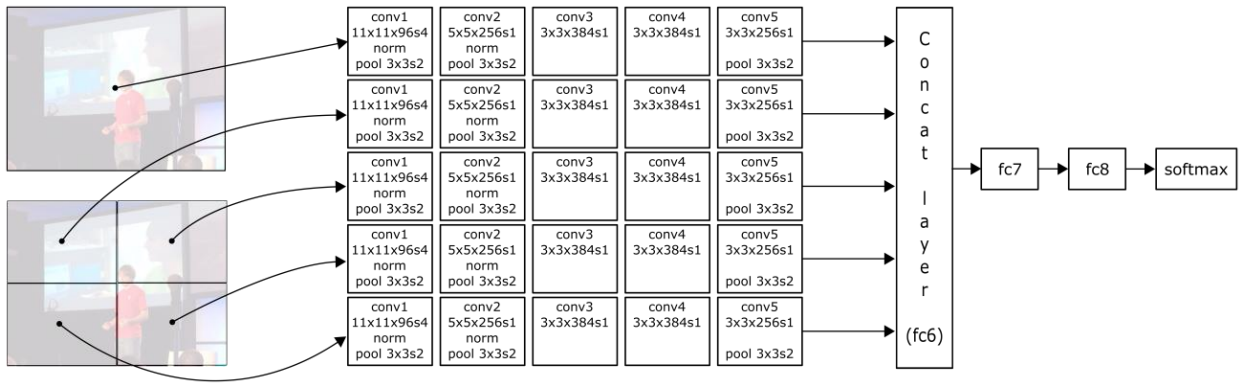

Fig. 1. The overview of our spatial pyramid CNN classifier based on Alex-Net CNN classifier. The convolutional streams are concatenated in layer fc6 or the first fully connected layer in Alex-Net CNN classifier.

|   | concert | graduation | meeting | mountaintrip | picnic | sea_holiday | ski_holiday | wedding |
|---|---|---|---|---|---|---|---|---|
| concert | 91.1 | 2.6 | 2.2 | 0.7 | 1.6 | 0.2 | 0.5 | 1.0 |
| graduation | 0.7 | 79.3 | 6.7 | 2.3 | 5.2 | 0.3 | 1.4 | 4.0 |
| meeting | 0.6 | 8.7 | 70.2 | 3.4 | 7.4 | 0.7 | 2.8 | 6.2 |
| mountaintrip | 0.0 | 0.2 | 0.4 | 94.6 | 1.6 | 0.8 | 2.0 | 0.3 |
| picnic | 0.2 | 6.5 | 6.1 | 15.2 | 61.0 | 1.6 | 3.2 | 6.3 |
| sea_holiday | 0.0 | 0.6 | 1.2 | 8.9 | 3.1 | 82.9 | 2.0 | 1.2 |
| ski_holiday | 0.0 | 1.1 | 1.7 | 16.3 | 2.3 | 1.1 | 76.4 | 1.0 |
| wedding | 0.3 | 4.7 | 5.1 | 3.4 | 5.2 | 0.7 | 1.8 | 78.8 |

Fig. 2. The confusion matrix of SP-CNN with USED-EiMM dataset with the average accuracy of 79.29%.

|   | concert | conference | exhibition | fashion | sport | protest | theater |
|---|---|---|---|---|---|---|---|
| concert | 82.4 | 1.4 | 1.4 | 2.1 | 0.5 | 1.1 | 11.2 |
| conference | 0.3 | 68.5 | 15.4 | 4.1 | 2.2 | 4.4 | 5.1 |
| exhibition | 0.2 | 14.5 | 59.1 | 9.5 | 3.2 | 6.2 | 7.2 |
| fashion | 0.3 | 3.3 | 7.7 | 74.1 | 2.6 | 3.2 | 8.9 |
| sport | 0.1 | 2.1 | 4.0 | 3.1 | 84.2 | 3.5 | 3.0 |
| protest | 0.4 | 6.9 | 10.1 | 6.3 | 5.6 | 63.3 | 7.5 |
| theater | 2.4 | 3.9 | 6.8 | 9.5 | 2.2 | 2.8 | 72.5 |

Fig. 3. The confusion matrix of SP-CNN with USED-SED dataset with the average accuracy of 72.0%.

*C. The Concatenated Layer*

The concatenated layer is the most important layer in the spatial pyramid CNN classifier because the last convolutional layer in Alex-Net CNN architecture has different resolution for each area of the spatial pyramid level. The concatenated layer works by flattening the convolution result from the previous layer in each stream and concatenated each flatten data into a big single column vector. The big single column vector is treated as input into fully connected layers. The total number of features that processed through fully connected layer is 21,760 features after concatenating all results of last convolutional layers.

III. THE RESULTS

The final classification decision of the image is taken directly from the classifier score. The video data is different and the final classification decision for the video data is the average accuracy of each frame.

The experiments using EiMM subset has an average accuracy of 79.29% and Fig. 2 shows the detail accuracy of each classes in the EiMM subset. Comparing with original Alex-Net CNN classifier as a baseline, our system outperforms the baseline in all classes except for meeting class with 15% better average accuracy. The experiments proved that SP-CNN classifier achieved very good accuracy and give the classifier chance to learn detail part of the image.

The average accuracy of the SP-CNN with SED subset is 72.0% and it's outperforms the Alex-Net CNN classifier baseline by 2%. Fig. 3 show the confusion matrix of the SP-CNN with SED subset in USED dataset and the lowest accuracy of the SP-CNN is exhibition class. The average accuracy is not increased so significant with the Alex-Net CNN classifier baseline but the SP-CNN classifier proved can learn the detail part of the image. As seen in Fig. 3, the SP-CNN seems to have a difficulty to learn event in conference, exhibition, and protest class which may have similar place, crowd, and maybe activities.

IV. CONCLUSION

We present the spatial pyramid convolutional neural network (CNN) to tackle the event detection in the image. The spatial pyramid CNN classifier work very well and achieve the average accuracy of 79.29% with EiMM dataset and 72.0% with SED dataset. The spatial pyramid configuration proved to successfully increase the accuracy of the system and ensured the classifier to extract the detail region of the image. Comparing with original dataset paper, our spatial pyramid CNN classifier outperforms the original Alex-Net by 15% with EiMM dataset and 2% with SED dataset.